\documentclass[conference]{IEEEtran}
\IEEEoverridecommandlockouts
\usepackage{amsmath,amssymb,amsfonts}
\usepackage{algorithmic}
\usepackage{braket}
\usepackage{graphicx}
\usepackage{textcomp}
\usepackage{xcolor}
\usepackage{graphicx}
\usepackage{cite}
\usepackage{epstopdf}
\usepackage{flushend}
\usepackage{url}
\usepackage{multirow}
\usepackage{lipsum}
\usepackage{filecontents}
\usepackage{multicol}
\usepackage{calc}
\usepackage{subfigure}
\usepackage{float}
\usepackage{stfloats}
\def\BibTeX{{\rm B\kern-.05em{\sc i\kern-.025em b}\kern-.08em
    T\kern-.1667em\lower.7ex\hbox{E}\kern-.125emX}}
\usepackage{booktabs,siunitx, array,threeparttable}
\restylefloat{table}

\begin{document}



 \title{Hybrid Quantum-Classical Machine Learning for Sentiment Analysis \\
}

\author{
Abu Kaisar Mohammad Masum, Anshul Maurya, Dhruthi Sridhar Murthy,\\
Pratibha, and Naveed Mahmud\\
\textit{Electrical Engineering and Computer Science}\\
\textit{Florida Institute of Technology}\\
\text{\{amasum2022, amaurya2022, dsridharmurt2022, pratibha2023\}@my.fit.edu}\\
\text{\{nmahmud\}@fit.edu}\\
[-2.8ex]
}

\maketitle

\begin{abstract} 
The collaboration between quantum computing and classical machine learning offers potential advantages in natural language processing, particularly in the sentiment analysis of human emotions and opinions expressed in large-scale datasets.
In this work, we propose a methodology for sentiment analysis using hybrid quantum-classical machine learning algorithms. We investigate quantum kernel approaches and variational quantum circuit-based classifiers and integrate them with classical dimension reduction techniques such as PCA and Haar wavelet transform. The proposed methodology is evaluated using two distinct datasets, based on English and Bengali languages. Experimental results show that after dimensionality reduction of the data, performance of the quantum-based hybrid algorithms were consistent and better than classical methods. 


\end{abstract}


\section{Introduction}

Sentiment analysis or opinion mining is the process of analyzing digital text to determine the emotions and opinions embedded in the text,
and helps shed light on important issues of human opinion \cite{Liu}.
By examining data from sources such as emails, tweets, chat transcripts, reviews, etc., sentiment analysis tools provide valuable insights into public opinions. It is a reliable method for predicting human behavior and has essential use cases such as improving customer service, brand monitoring, and product market research. 

In recent years, researchers have demonstrated the efficiency of machine learning techniques for the correct classification of sentiments \cite{Taherdoost}. 
However, due to the computational limitations of classical machine learning \cite{Huang}, researchers are investigating the possibilities of newer, more efficient paradigms of computing, such as quantum computing. Quantum computers can take advantage of quantum mechanical effects such as superposition and entanglement to provide exponential speedup in specific tasks \cite{nielsen2010quantum} compared to state-of-the-art classical computers. The rapid advancement of noisy 
intermediate-scale (NISQ) quantum hardware has driven research into quantum algorithm development for a variety of multidisciplinary fields such as Quantum Machine Learning (QML) \cite{Biamonte}.
Integrating machine learning with quantum computing can lead to greater benefits such as improved text classification \cite{Shah}. 
Recent studies have introduced a quantum support vector machine (QSVM) for conventional data, to improve the performance of the traditional support vector machine (SVM) \cite{Zhang}. 
A QSVM uses the quantum kernel approach \cite{Park}, which is essentially a hybrid approach that uses a classical SVM as the classifier. 
A type of quantum classifier known as the Variational Quantum Classifier (VQC) \cite{enhanced_Feature_mapping} has also been introduced that uses parameterized quantum circuits.

The limited scalability of current NISQ devices makes it challenging to encode high-dimensional data for QML tasks. Dimension reduction techniques can be used to decrease the data dimensionality by identifying the most pertinent features from data and/or demoting the input number of data points. 
Principal component analysis, also referred to as PCA, is a widely known approach for obtaining the principal features of the data \cite{Cao}. 
Another data compression technique widely used in signal processing is the Wavelet Transform, which preserves the spatial locality of data, and compared to other transforms, also provides better computation speeds \cite{El-Araby}.

In this work, we investigate the use of quantum machine learning techniques for sentiment analysis of textual data, and we propose dimension reduction techniques to reduce the dimensionality of the textual data. We present a methodology combining conventional Natural Language Processing (NLP) techniques with QML models such as QSVM and VQC. The proposed methodology also integrates techniques such as PCA and Wavelet transform to reduce data dimensionality, making it convenient to fit the classical data onto the quantum models and reducing the algorithm execution time. 

Our methodology is evaluated by employing textual data from two languages: English and Bengali. NLP for English text has been explored, for instance, Zhang et al. \cite{Zhang_W} investigated English text classification.
The structure of the Bengali text is complex and different from other languages, especially in \textit{pre-processing} and \textit{word embedding} \cite{Abujar} phases.
This work presents efficient models for \textit{pre-processing} and \textit{word embedding} in support of both Bengali and English language-based text. 
In addition, we investigate the efficiency of quantum models for feature mapping and classification, and compare their performance with classical methods, when applied for sentiment analysis.
An important contribution of this research is the use of Wavelet transform for dimensionality reduction of textual data. 
Wavelet-based feature extraction enables a meaningful representation of the data with reduced dimensionality, and combined with hybrid quantum-classical ML models, leads to improved sentiment analysis.

\section{Related Work}
In this section, we discuss the state-of-the-art methods (quantum or hybrid) that have been proposed for sentiment analysis. Ganguly et al. used the lambeq toolkit for sentiment analysis \cite{Ganguly}. They performed noisy simulations on an IBM backend, achieving 83.33\% accuracy. However, they used a very small dataset (130 sentences), and their test accuracy could be improved for noisy quantum simulation by performing more iterations. 
Morales et al. \cite{Morales} proposed a framework for quantum natural language processing (QNLP) and implemented their work on near-term quantum computers. 
In their work, they leveraged variational quantum circuits in QNLP models.
Further research and empirical studies are necessary to address the challenges involved and understand the potential benefits and limitations of applying quantum computing in natural language processing. 
In their paper, Meichanetzidis et al. \cite{Meichanetzidis} presented a pipeline for QNLP. The framework used is compositional distributional semantics (DisCoCat), which extends the compositional structure of group grammars. 
Overall, the paper provides an overview of the proposed QNLP pipeline, but it lacks empirical evidence, comparative analysis, and in-depth discussions on the feasibility and limitations of the approach. 

Existing quantum sentiment analysis models primarily rely on rule-based techniques to process and analyze sentiment data. These models utilize quantum machine learning algorithms on a small amount of text data to extract sentiment. On the other hand, our hybrid quantum-classical sentiment analysis model follows automatic approaches for combining both quantum and classical computing techniques. Instead of solely relying on quantum algorithms, we leverage the strengths of classical algorithms alongside quantum algorithms to enhance sentiment analysis performance. 


\section{Proposed Methodology}


\begin{figure*}[hbt!]
 \center
  \includegraphics[width=16cm,height=5cm]{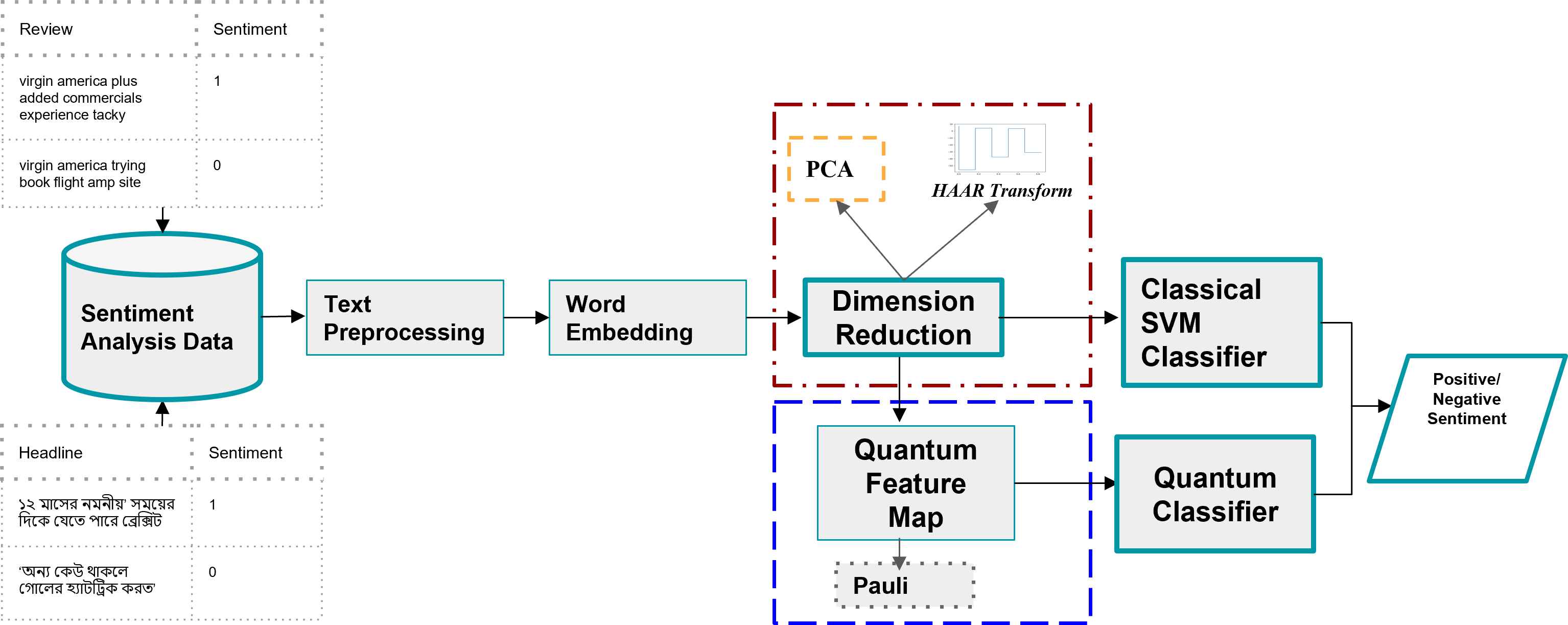}
  \caption{Workflow for the proposed hybrid quantum-classical methodology for sentiment analysis.}
\end{figure*}

In this work, we present a methodology for performing sentiment analysis of languages by combining classical and quantum machine learning techniques. At first, the input data undergoes pre-processing, then a word embedding algorithm converts pre-processed data to a word vector. After that, we apply dimension reduction algorithms to reduce data dimensionality. The consequent data is then fed to either classical SVM models, or to quantum models, for training and classification of sentiment. 
The workflow of the proposed methodology is shown in Fig. 1 and we discuss its further details in the rest of this section.

\subsection{Sentiment Analysis Data}
In this work, we have used two datasets for evaluating the methodology. The Bengali dataset \cite{Bengalidata} consists of news data where each line of text reflects human opinion. Scripts were used to collect the data from Bengali news portals. Furthermore, positive and negative sentiments have been separated into two categories of data. The dataset is structured as two columns of Bengali text and their correlated class and contains a total of 1619 data points and 65 distinct features. Each line of text contains vital words that represent the text context.
Another dataset from \textit{``Twitter US Airline Sentiment"}\cite{Tiwari} has been used to measure the performance of the implemented model. The dataset has a total of 728 data points and 2027 unique features. For the Bengali and Twitter datasets, two separate word vector models subsequently converted them into two sets of feature vectors.

\subsection{Text Pre-processing}
Pre-processing Bengali text is distinct from other languages. Bengali words have a complex format. 
Therefore, additional study is needed to develop rule-based grammar systems for the Bengali language and we primarily focused on basic text preparation and automated systems. 

We have implemented a function for pre-processing the Bengali text which includes techniques such as sentence-to-word \cite{Grefenstette}, regular expression \cite{Cui}, and stop-word removal \cite{Silva}. The text is processed from raw data using regular expression and then each sentence is split into words. At the same time, the stop words are removed from the list of stop words. Consequently, the processed text is sent to the next stage, i.e., word embedding, see Fig. 1. Fig. 2 gives a model diagram for the text pre-processing phase for Bengali text.  For the Twitter dataset, we have applied available pre-processing approaches like tokenizer \cite{Grefenstette}. To facilitate the training of data, the output column must be converted into binary labels. To achieve this, a label encoder based on scikit-learn is utilized to transform the output column into binary labels.

\begin{figure}[H]
    \includegraphics[width=7cm, height=3cm]{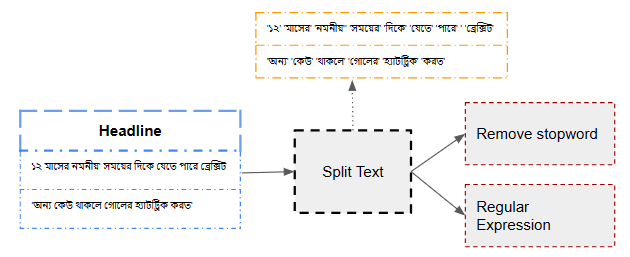}
    \caption{Bengali text pre-processing method for sentiment analysis.}\label{fig:a}
\end{figure}
\subsection{Word Embedding}
We have leveraged the count vectorization technique \cite{Wang} as a method of text data representation. 
By definition, count vectorization is a type of Boolean concept formulation in document labeling terminology, where documents can be represented via a set of words.
In count vectorization, the dimension of the word vectors is determined by the length of the vocabulary. Whenever a new word occurs, the number of words and dimensions are simultaneously incremented by one. A counting vectorizer's fundamental objective is to comprehend and accommodate every single word in the lexicon. Fig. 3 shows a model of the count vectorizer used in our approach. A document phrase matrix has been constructed on the basis of the new lexicon that was developed. In our approach, every data row comprises a collection of words that signifies specific information. All the words within these data rows convey either negative or positive sentiments \cite{Turki}. 

\begin{figure}[H]
    \includegraphics[width=9cm]{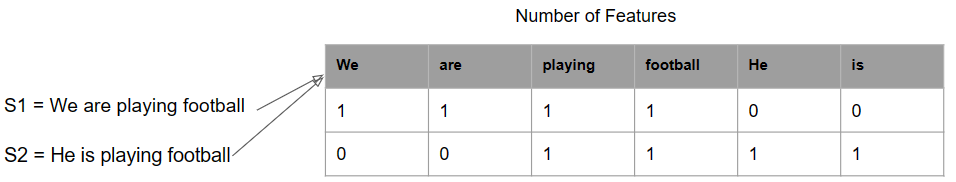}
    \caption{Count vectorization approach for word embedding.}\label{fig:a}
  \end{figure}

\subsection{Dimension Reduction}
For the proposed sentiment analysis methodology, we integrate two different techniques for data dimensionality reduction: Principal Component Analysis for dataset feature reduction and Haar Wavelet Transform for data compression.
\paragraph{Principal Component Analysis}
In our PCA model, linear arrays are used at the beginning to determine the major vocabulary elements from the text datasets. The first principle component shows the initial set of data with the highest variance, while the second principle component shows the rest of the dataset with the most variation. The process continues sequentially, with the subsequent principal component representing the set of data from the rest of the dataset having the highest level of variance \cite{Uguz}. The most influential phrases are subsequently put to feature extraction procedures. $M$ represents the overall count of principal components, while $P$ denotes the substantial percentage of principal elements present in each principal component. However, $P$ is commonly understood as the total count of principal elements present in the $M$-dimensional spaces. This represents the highest percentage of variable value pairs. 
The PCA principles provide clear evidence that $P \leq M $, as derived from the rules. This supports the data dimension reduction approach for our datasets. Nevertheless, PCA is the most useful data compression method for classical systems where the method depends on the corresponding vector dimension of the text.

\paragraph{Haar Wavelet Transformation}
Haar transform holds significant prominence in the realms of signal and image processing \cite{fastPatternMatching}. Haar transform makes it possible to decrease the number of data points while preserving the locality of the data and the relationships between neighboring values. The motivation behind utilizing this compression technique in our approach is to reduce the dimensionality of the data, leading to faster quantum circuit execution time in the consequent stages of our proposed methodology. The inverse of the Haar transform \cite{HaarTransformPaper} can also be used to get back the original data points after dimension reduction.

The Haar transform can be generalized as $n$-Dimensional ($n$-D) Haar transform, where $n$ is the dimensionality of the data. Detailed algorithms for 1-D and 2-D Haar Transform can be found in \cite{HaarTransformPaper}. 
In this work, we employed a $1$-D discrete Haar Transform due to the discrete nature of the dataset and to reduce the dimensionality of the dataset.
The $1$-D Haar Transform can be performed as a multi-level, decomposable operation \cite{HaarTransformPaper}. Each level of operation involves dividing the set of data points into two distinct non-overlapping segments. This division is achieved by computing the average and difference between neighboring values. The resulting difference values are generally close to zero or significantly smaller in comparison to the averaging values. Consequently, these negligible difference values can be safely discarded. By discarding them, the transformed dataset retains only the high-amplitude values, effectively reducing the dimension(s) of the original dataset by exactly half. For $l$ decomposition levels, the dimension(s) will be reduced by $2^l$.

\subsection{Classical SVM Classifier}
SVM is a classical machine learning method that allocates labels to data objects after training on example data \cite{Noble}. 
For example, an SVM can be used for negative and positive sentiment analysis on a text-based dataset. The SVM method works by statistical classification
where a separate hyperplane and a maximum number of margins between two separated objects work as base points. The linear SVM classifier is considered the most fundamental and quickest SVM, supposing a straight-line separation between classes. It was defined as the most effective technique for text categorization \cite{Xia}. 
Our sentiment analysis methodology employs a traditional SVM classifier to accommodate each data point within $n$-dimensional spaces. Within these spaces, the feature values are extracted based on the corresponding data point's coordinates.

\subsection{Quantum Feature Mapping}
Quantum computers harness the power of an exponentially large quantum state space, offering great potential for handling large datasets and mapping them to higher dimensions. Quantum feature mapping serves as a means to non-linearly map classical data onto quantum states, represented as vectors in Hilbert space, $\ket{f(x)}\bra{f(x)}$; where $f$ is a data mapping function and $x$ is the feature data point \cite{enhanced_Feature_mapping}. 

A quantum circuit that is classically hard to simulate is important to gain the quantum advantage, as it provides a mapping of the data that is hard to compute classically. Entanglements between qubits in such circuits, take account of nonlinear interactions between features \cite{enhanced_Feature_mapping}. In our work, we have employed a generalized second-order Pauli feature mapping, discussed further:



\paragraph{Pauli feature map}
A generalized method for feature mapping that utilizes Pauli basis gate set \cite{nielsen2010quantum}. \textcolor{black}{Based on the mapping function $f$, we design a unitary} utilizing Pauli gate set \{X, Y, and Z\}, and an entanglement scheme which is typically either \textit{linear} or \textit{full} entanglement. The full entanglement feature map circuit will have full connectivity between qubits, accounting for $n(n-1)\//2$ feature interactions, where $n$ is the number of qubits in the circuit. Linear entanglement scheme accounts for interactions with only adjacent qubits. Thus, the full entanglement feature map has a higher circuit depth than the linear entanglement scheme. A classically hard to simulate \cite{hardToSimulate} Pauli feature map unitary for the gate pair of $Z$ and $XX$ is shown in equation (\ref{eq:unitaryEquation}), accounting for two-qubit interaction.

 


\begin{equation} \label{eq:unitaryEquation}
\resizebox{0.42\textwidth}{!}{$
\begin{aligned}
U_{f(x)} = \left(\exp\left(i \sum_{q_j,q_k}f_{q_j,q_k} (x)X_{q_j}\otimes X_{q_k}\right) \times \exp\left(i\sum_{q_j}f_{q_j}(x)Z_{q_j} \right)H^{\otimes n}\right)^r
\end{aligned}
$}
\end{equation}

Here $X$ and $H$ are conventional NOT and Hadamard gates respectively, $XX$ is the tensor of two $X$ gates, and $r$ is the number of circuit repetitions. The unitary $U_{f(x)}$ will be applied to $\ket{0}$, the initial state of the qubits.
\textcolor{black}{The $H$ gate puts the circuit in the superposition state and phase gate (rotation gate in Pauli basis) manipulates the superposition state based on the chosen function map $f(x)$. 
In \eqref{eq:unitaryEquation}, the exponent involving the tensor operation ($X\otimes X$) generates the entangled portion of the feature map circuit, and the remaining exponent terms, contribute to the rest of the non-entangled part of the circuit.}

\subsection{Quantum Classifiers}

\paragraph{SVM classifier with quantum kernel}
SVM can handle linearly separable data points but non-linearly separable data can be classified by employing kernel functions like linear, polynomial, and sigmoid functions, which maps the data to higher dimensions. \textcolor{black}
The expression to calculate the quantum kernel matrix elements is $K_{i,j} = \mod{(\bra{f(x_i)}^{\dagger}\ket{f(x_j)})}^2$.
\textcolor{black}{$K_{i,j}$ is the measure of the distance between each data point with every other in the dataset.}
After computing the kernel matrix, it can be utilized as input for any standard classical machine-learning model to train a hyperplane.

\paragraph{Variational Quantum Classifier (VQC)}
A quantum algorithm for classification known as variational quantum classifier (VQC) was introduced in \cite{enhanced_Feature_mapping}.
In the VQC algorithm, a variable parameterized quantum circuit, also known as ansatz, is used in conjunction with quantum feature maps for classification.

In VQC, the output of the quantum feature map is passed through the ansatz, which typically consists of parameterized rotation gates and CNOT gates\cite{enhanced_Feature_mapping}. Consequently, measurement of the circuit produces a classical bitstring that is mapped to a binary label, and then matched with the corresponding binary label of the encoded feature set in the feature map. Later this mapping gets utilized by a classical system which tunes the rotation parameters by 
optimizing the cost function $\bra{f(x)}U^{\dagger}\hat{M}U\ket{f(x)} \geq \Delta$ \cite{enhanced_Feature_mapping},
where $U$ is the random unitary which initializes the parameters of ansatz and $\hat{M}$ is the measurement typically in a Z basis. The value of $\Delta$ decides the separation between the labels. In our work, we have used a real amplitude circuit as an ansatz with linear entanglement scheme. The real amplitude circuit is based on the $Y$-rotation gate ($R_y$) and only affects the real components of the quantum state. 

\begin{table*}[hbt!]
\centering
\includegraphics[width=12cm]{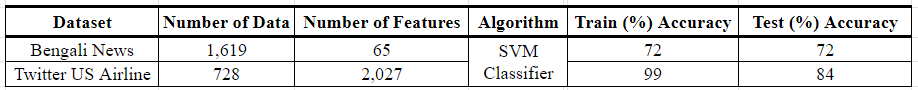}
\caption{Performance of Classical SVM for Bengali and Twitter Datasets.}
\label{table:Table 1}
\end{table*}

\begin{table*}[hbt!]
\centering
\includegraphics[width=14cm]{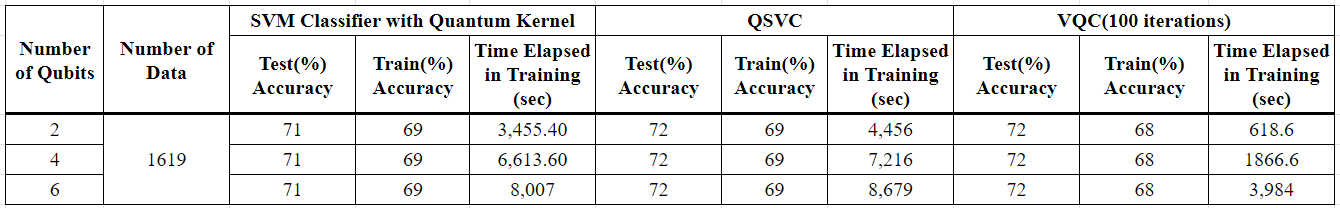}
\caption{Performance comparison of quantum-classical methods using Pauli feature map and PCA on Bengali Dataset.}
\label{table:table 3}
\end{table*}

\begin{table*}[hbt!]
\centering
\includegraphics[width=14cm]{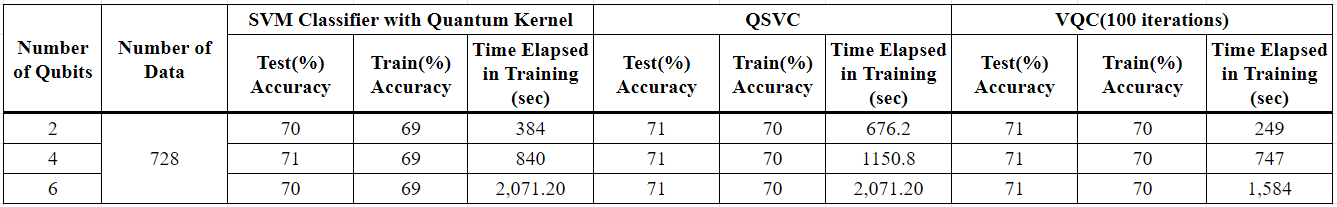}
\caption{Performance comparison of quantum-classical methods using Pauli feature map and PCA on Twitter Dataset.}
\label{table:table 5}
\end{table*}
\section{Experiment Results And Analysis}

\begin{table*}[hbt!]
\centering
\includegraphics[width=14cm]{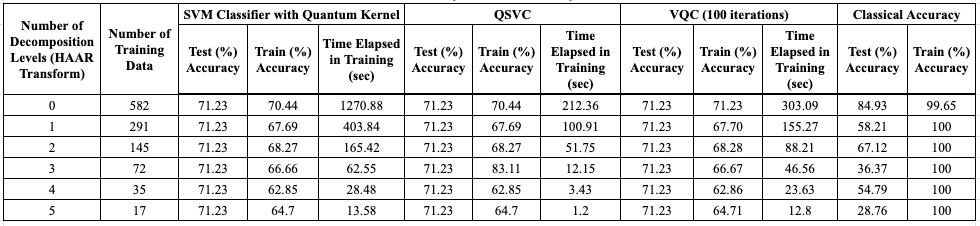}
\caption{Performance comparison using Pauli feature Map and PCA + Haar compression on Twitter Dataset.}\label{table:twitterPauli}
\end{table*}

\begin{table*}[hbt!]
\centering
\includegraphics[width=14cm]{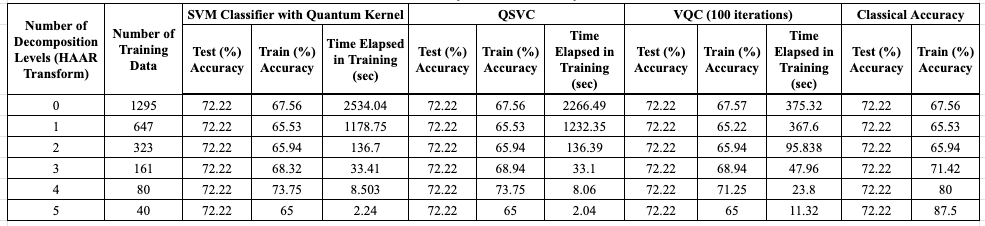}
\caption{Performance comparison using Pauli feature Map and PCA + Haar compression on Bengali Dataset.}\label{table:banglaPauli}
\end{table*}
The experiments to evaluate the proposed hybrid methodology were conducted using qiskit \cite{qiskit}. 
The experiments were performed on the noisy \textit{qasm\_simulator} \cite{QSVM} backend from IBM. 
For evaluation we used two distinct datasets containing English and Bengali text data. We classified text data into two classes, positive and negative based on the sentiment of the data.

To establish a baseline, we first performed sentiment analysis on the datasets using fully-classical SVM models without any dimension reduction. The obtained results are presented in Table \ref{table:Table 1}.
For the Bengali dataset, the classical accuracy on Test data is 72\%, while for the English dataset, the accuracy is 84\%. 
In this work, we have used namely the following quantum-classical methods, the SVM classifier with quantum kernel and the Variational Quantum Classifier (VQC), which is a parameterized quantum circuit classifier.
The VQC circuit parameters were optimized using ADAM optimizer \cite{adamOptim}.
For comparison, we used a third method called the Quantum Support Vector Classifier (QSVC), which is another variant of SVM classifier with a quantum kernel and is directly imported from qiskit's library. The QSVC is a pre-built qiskit function which takes a feature map input and trains an SVM classifier. 


We implemented quantum circuit-based, classically-hard-to-simulate feature mapping by synthesizing \eqref{eq:unitaryEquation}. 
For our experiments, the data mapping function $f$ used is: 
    \begin{equation}\label{eq:f-2}\begin{aligned}
         f_{q_j,q_k}(x) = \left(\pi-x[j]\right)\left(\pi-x[k]\right)
    \end{aligned}\end{equation}
    \begin{equation}\label{eq:f-1}\begin{aligned}
         f_{q_m}(x) = x[m]
    \end{aligned}\end{equation}
    

\noindent Equation \eqref{eq:f-2}, will be utilized for an entangled qubit pair, $q_j$ and $q_k$, and $x$ represents the feature data point, while \eqref{eq:f-1} is applicable for non-entangled qubits. The entanglement arrangement we used in the experiments is linear entanglement.



In Table \ref{table:table 3}, we present the performance of the implemented quantum-classical methods when using Pauli feature maps. We evaluate their effectiveness on the Bengali dataset reduced to 2, 4, and 6 feature sets after PCA. The test accuracy consistently falls within the range of 71\% to 72\% for all feature sets.

We also measured the training times for each technique, and VQC required the least training time for each feature set compared to other algorithms, see Table \ref{table:table 3}. The results in Table \ref{table:table 5} for the Twitter dataset using  Pauli feature maps, respectively demonstrate consistent accuracy for both training (69\%-71\%) and testing (70\%-71\%). Notably, when training the feature sets, the QSVC and VQC algorithms performed with higher test and train accuracies compared to the SVM classifier with quantum kernel, see Table \ref{table:table 5}. 
Furthermore, VQC exhibits shorter training times compared to other algorithms when applied to this dataset also.



With the number of features reduced to two using PCA, we applied Haar transform to further compress the number of data points in the datasets. This significantly reduced the training time of the subsequent quantum methods. Up to five levels of Haar decomposition were performed. In addition to the quantum-classical methods, we also evaluate a classical SVM classifer for accuracy comparison with the quantum-classical methods.
Table \ref{table:twitterPauli} show that only one level of decomposition reduced the classical SVM test accuracy to about 58\% with  further degradation at higher decomposition levels. On the other hand, the quantum-classical algorithms provide better and steady classification accuracy of 71.23\%. This implies that the Haar transform in combination with the quantum-classical methods is more effective than when applied for the classical SVM.
The VQC algorithm takes less time to train than other methods on average and maintains consistent classification accuracy up to the maximum number of decomposition levels applied. 
For the Bengali dataset, see Table \ref{table:banglaPauli}, the classical accuracy is constant for up to the maximum decomposition level because the dataset used in the experiment is balanced. Quantum-classical algorithms for the Bengali dataset after dimensionality reduction show the same level of performance as classical systems without dimension reduction, achieving an accuracy of 72.22\%, see Tables \ref{table:Table 1}, and \ref{table:banglaPauli}. This implies that for the Bengali dataset, our quantum-classical methods in combination with dimension reduction was highly effective, leading to efficient sentiment analysis on a large dataset. 

\section{Conclusion and Future Work}
Our research demonstrates the potential of quantum-classical hybrid algorithms in sentiment analysis of Bengali and English datasets. 
We also highlighted the importance of considering language-specific characteristics, particularly in Bengali sentiment analysis. 
We investigated hybrid quantum-classical algorithms that included quantum feature maps and quantum classifiers.
Moreover, we integrated dimension reduction techniques to facilitate encoding classical data on to the quantum models. 
Moving forward, there are several areas for future work in quantum-classical hybrid sentiment analysis research. Firstly, expanding the analysis to include other languages would provide a more comprehensive understanding of the effectiveness of hybrid techniques across different linguistic contexts. Additionally, exploring error mitigation and noise correction, and the translation of quantum sentiment analysis into practical quantum applications are important areas for further investigation.

\bibliographystyle{unsrt}
\bibliography{references}

\vspace{12pt}
\end{document}